\pdfoutput=1

\documentclass[11pt]{article}

\usepackage{acl}

\usepackage{times}
\usepackage{latexsym}

\usepackage[T1]{fontenc}

\usepackage[utf8]{inputenc}

\usepackage{microtype}

\usepackage{graphicx}
\usepackage{bm}
\usepackage{amsmath}
\usepackage{amsfonts}
\usepackage{bbm}

\usepackage{helvet}
\usepackage{courier}
\usepackage{multirow}
\usepackage{bm}
\usepackage{amsmath,amssymb}
\usepackage{mdwlist}

\usepackage{relsize}
\usepackage{amsfonts}
\usepackage{url}
\usepackage{graphicx}
\usepackage{subfigure}
\usepackage{caption}
\usepackage{hhline}
\usepackage{color}
\usepackage{colortbl}
\usepackage{booktabs}
\usepackage{tabularx}
\usepackage{makecell}
\usepackage[linesnumbered,boxed,ruled,commentsnumbered]{algorithm2e}
\usepackage{CJKutf8}  
\usepackage{lipsum}
\title{Advancing Translation Preference Modeling with RLHF: A Step Towards Cost-Effective Solution}

\author{Nuo Xu$^{1}$\footnotemark[1]\;,\ \ Jun Zhao$^{1}$\footnotemark[1]\, \footnotemark[2]\;,\ \ Can Zu$^{1}$, \ \ Sixian Li$^{1}$,\ \ Lu Chen$^{1}$,\ \ Zhihao Zhang$^{1}$,\ \ Rui Zheng$^{1}$,\\\ \ \textbf{Shihan Dou}$^{1}$,\ \ \textbf{Wenjuan Qin}$^{3}$, \textbf{Tao Gui}$^{2}$\footnotemark[2]\;,\ \ \textbf{Qi Zhang}$^{1}$\footnotemark[2]\;,\ \ \textbf{Xuanjing Huang}$^{1}$\\
  $^1$School of Computer Science, Fudan University\\
  $^2$Institute of Modern Languages and Linguistics, Fudan University\\
  $^3$ College of Foreign Languages and Literature, Fudan University\\
  \texttt{xun22@m.fudan.edu.cn,\{zhaoj19,qz,tgui\}@fudan.edu.cn}}
\begin{document}
\maketitle
\renewcommand{\thefootnote}{\fnsymbol{footnote}}
\footnotetext[1]{Equal Contributions.}
\footnotetext[2]{Corresponding authors.}
\begin{abstract}
Faithfulness, expressiveness, and elegance is the constant pursuit in machine translation. However, traditional metrics like \textit{BLEU} do not strictly align with human preference of translation quality. In this paper, we explore leveraging reinforcement learning with human feedback (\textit{RLHF}) to improve translation quality. It is non-trivial to collect a large high-quality dataset of human comparisons between translations, especially for low-resource languages. To address this issue, we propose a cost-effective preference learning strategy, optimizing reward models by distinguishing between human and machine translations. In this manner, the reward model learns the deficiencies of machine translation compared to human and guides subsequent improvements in machine translation. Experimental results demonstrate that \textit{RLHF} can effectively enhance translation quality and this improvement benefits other translation directions not trained with \textit{RLHF}. Further analysis indicates that the model's language capabilities play a crucial role in preference learning. A reward model with strong language capabilities can more sensitively learn the subtle differences in translation quality and align better with real human translation preferences.
\end{abstract}

\section{Introduction}
As a crucial technology facilitating communication between disparate languages and cultures, machine translation has long garnered significant attention from both academia and industry \cite{yang2020survey}. Recently, the emergence of large language models (LLMs) has propelled the field to new frontiers \cite{yang2023bigtranslate,zhu2023multilingual,jiao2023chatgpt,hendy2023good}. Pre-training on massive monolingual datasets has alleviated the reliance on extensive parallel corpora while enhancing translation quality \cite{xu2024paradigm}. 


To enhance the translation capabilities of models, much of the research works have adopted one of two optimization objectives: one is through supervised fine-tuning of translation models to maximize the log probability of human translations \cite{yang2023bigtranslate,xu2024paradigm}; the other is through the techniques like reinforcement learning, directly optimizing the similarity score (e.g., \textit{BLEU} score \cite{10.3115/1073083.1073135}) between model outputs and human translations \cite{ranzato2016sequence,wu-etal-2018-study,wieting-etal-2019-beyond}. Although both approaches have generally performed well, the objectives they optimize for are not fully aligned with human's preferences for translation faithfulness, expressiveness and elegance \cite{COMET, summarize}. 

Fortunately, reinforcement learning from human feedback (RLHF) has been shown to be effective in aligning model behavior with human societal values \cite{ouyang2022training,bai2022training}. This process integrates reward modeling, where human annotators rank different responses from models based on their preferences, and then normalizes model behavior through a reinforcement learning (RL) phase. However, it is non-trivial to collect a large high-quality preference dataset. Firstly, preference data often comes with noise and ambiguity, as there is low consistency among different human annotators \cite{wang2024secrets}. More importantly, preference data annotation for translation tasks places higher demands on annotators' linguistic capabilities, a challenge particularly pronounced in low-resource languages.

This paper explores improving translation quality through RLHF and proposes a cost-effective preference learning strategy. We avoid the need to construct expensive preference datasets and instead leverage the inductive bias that \textit{high-quality human translations are superior to machine-generated translations.} The reward model learns human translation preferences by comparing the quality difference between the two, and subsequently guides the improvement of machine translation quality. To collect such high-quality human translations, we align books with multilingual versions. Our motivation for choosing books as the data source is as follows: 1) the original text is authored by writers and the target language is translated by professional translators, ensuring the quality of both texts; 2) compared to web text, book text typically contains more complex language structures, which is particularly beneficial for learning translation preferences; 3) aligning book text does not require as high a level of linguistic capabilities from annotators and can be assisted with external tools \cite{wang2023guofeng}. The experimental results indicate that the reward model effectively learns human translation preferences, and the translation quality of the model is significantly improved. 

The main contributions of this paper are as follows: 1) We explore the use of RLHF to improve machine translation quality and propose a cost-effective preference learning strategy that avoids the need for expensive preference data construction; 2) Our experimental results demonstrate that RLHF can improve translation quality, and this improvement can be transferred to target languages not trained with RLHF; 3) Further analysis shows that reward models with strong language capabilities can more sensitively learn differences in translation quality and have stronger resistance to noise in the data.

\section{Related works}
\subsection{Reinforcement Learning from Human Feedback}
In recent years, research applying RLHF techniques to tasks involving LLMs has significantly increased ~\citep{ouyang2022training,llama2}, aiming to align the behavior of these models more closely with human preferences. 
For instance, \citet{summarize} employ this technique to enhance the quality of summaries, while \citet{bai2022training} utilize it to enable the model to generate responses that are more harmless and useful. 

These technique follows a systematic approach: firstly, collect task-specific human preference data. Then, use this data to train a reward model, which acts as a proxy for human preferences. During reinforcement learning, this reward model provides signals to guide model training.
However, collecting human preference data is non-trivial, time-consuming, and labor-intensive, often requiring high demands on annotators and plagued by inconsistencies in annotation standards among them.~\cite{bai2022training,rlhf-problem,wang2024secrets}


\subsection{Human-like Alignment in Translation}
Achieving human-level machine translation has long been a research goal, receiving ongoing attention. ~\cite{microsoft,DBLP:journals/corr/WuSCLNMKCGMKSJL16,DBLP:conf/emnlp/LaubliS018} Recent years, some studies have focused on improving the quality of machine translation through human feedback and alignment techniques. ~\citet{ebay} gather implicit task-based feedback, enhancing individual word translations and automatic evaluation measures.
~\citet{parrot} 
employs contrastive instruction and error-guided  instruction to align LLMs with human feedback.
~\citet{qe} attempt to leverage the quality estimation model as the reward model to predict human preference feedback.

Considering the methods above, the scarcity of human-preference data in translation has long been a bottleneck. Our approach differs, creatively utilizing meticulously translated human data as readily available preference data.

\section{Improving Translation with RLHF}

\begin{figure*}[t]
    \centering
        \includegraphics[width=\linewidth]{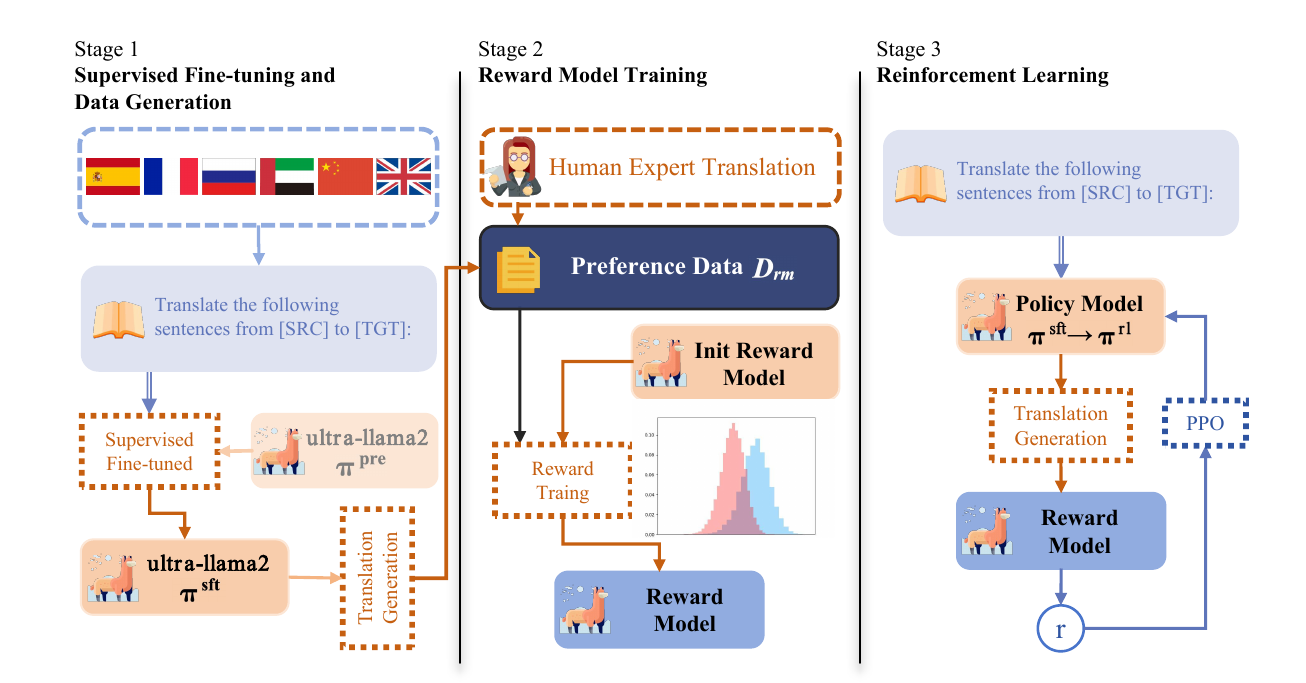}
        \caption{An Overview of Modeling Translation Preferences using RLHF; To achieve cost-effective preference learning, we optimize the reward model in the second step by contrasting the deficiencies of SFT model translations with human expert translations, thus avoiding the expensive labeling of preference data.}
        \label{fig:model}
    \end{figure*}

To build a translation model that aligns with human translation preferences, we start with a generic pre-trained language model $\pi^{\text{pre}}$ (such as LLaMA \cite{touvron2023llama}), and follow the pipeline of the following three steps: 1) Supervised fine-tuning of $\pi^{\text{pre}}$ on parallel corpora yields a model $\pi^{\text{sft}}$ with basic translation capabilities; 2) Training a reward model $r$ on preference dataset $\mathcal{D}_{\text{rm}}$, which assigns high reward scores to translations that adhere to human preference; 3) Utilizing $r$ as a proxy for human preferences, enhancing the translation quality of the model through reinforcement learning.
\subsection{Supervised Fine-tuning to Acquire Basic Translation Capabilities}
Given a parallel corpus $\mathcal{D}_{\text{sft}}=\{(x^{(i)},y^{(i)})\}_{i=1,..,n}$, where $x_i$ represents the source-language text and $y_i$ represents the corresponding reference translation, we utilize a fixed prompt template $\mathcal{I}$ and construct the training data as follows:

\centerline{$\mathcal{I}=$``Translate this from [SRC] to [TGT]:}
\centerline{[SRC]: <$x$> [TGT]: <$y$>''}
\noindent where, 'SRC' and 'TGT' respectively represent the names of the source language and the target language. The translation model $\pi^{\text{sft}}$ is optimized via the negative log-likelihood loss on parallel corpus $\mathcal{D}^{\text{sft}}$ as follows:
\begin{equation}
\mathcal{L}_{NLL}=-\mathbb{E}_{(x,y)\sim\mathcal{D}^{\text{sft}}}\log \pi^{\text{sft}}(y|x,\mathcal{I}),
\end{equation}
The translation model $\pi^{\text{sft}}$ acquired basic translation capabilities by maximizing the probability of reference translations.

\subsection{Modeling Translation Preferences}

To accurately model human preferences, high-quality preference data is crucial. A common practice used for modeling human value preferences is to prompt the model to generate two different outputs $(y1, y2)\sim \pi^{\text{sft}}(\cdot|x)$ in response to a query $x$ and then require annotators to choose their preferred one, i.e., $y_w > y_l$. $y_w$ and $y_l$ denote the chosen and rejected response, respectively. However, constructing a large preference dataset for translation tasks requires annotators who are experts/native speaker in the specific languages, which greatly increases the annotation cost. For low-resource languages, finding a sufficient number of qualified annotators may even be impractical.

Unlike the aforementioned approach, we instead leverage the induction bias of `\textit{high-quality human translation is superior to machine-generated translation}' to collect preference data at a lower cost. These high-quality human translations are sourced from book data. Our motivation for selecting this data source is as follows: 1) Books' original texts and their translated versions are completed by authors and professional translators, ensuring high text quality; 2) Book corpora contain more complex language structures compared to web text, which is highly beneficial for preference learning; 3) Aligning book data requires less stringent language proficiency from annotators and can be aided by external tools. 

We optimize our reward model $r$ by contrasting the differences between high-quality human translation and machine translation:
\begin{equation}
    \mathcal{L}(r)=-\mathbb{E}_{(x,y_w,y_l)\sim\mathcal{D}_{\text{rm}}}[log\sigma(r(x,y_w)-r(x,y_l))],
\end{equation}
where $x$ represents the source language sentence, while $y_w$ and $y_l$ respectively denote a high-quality human translation and a machine-generated translation, and $\mathcal{D}_{\text{rm}}=\{(x^{(i)},y^{(i)}_w,y^{(i)}_l)\}_{i=1,..,N}$ is the preference dataset.

\subsection{Improving Translation via RL Fine-tuning}
During the Reinforcement Learning (RL) phase, we employ the acquired reward function to furnish feedback to the language model. Specifically, we refine the policy model $\pi^{rl}$ to optimize the following reward objective:

\begin{equation}
    r_{total}=r(x,y)-\eta KL(\pi^{\text{rl}}(y|x)||\pi^{\text{sft}}(y|x)),
\end{equation}

where $\eta$ represents a coefficient regulating the extent of the KL penalty. The KL divergence component serves two main purposes within this framework. Firstly, it functions as an entropy bonus, maintaining diversity in generation and averting the collapse into singular high-reward responses \cite{DBLP:journals/corr/abs-1907-00456}. Secondly, it ensures that the output of the RL policy remains within a distribution where the reward model accurately reflects the performance, thereby preventing significant deviations.

\section{Experimental Setup}
\subsection{Training Data Collection}
\label{sec:data_collection}
We collect and utilize translation training data from three different sources. The detailed information of these datasets can be found in table ~\ref{mt-datasets}.


\begin{table*}
\centering
\resizebox{\linewidth}{!}{
\begin{tabular}{cccc}
\toprule 
\textbf{Name of the dataset} & \textbf{Translation direction} & \textbf{Granularity} & \textbf{Training Samples}\\
\hline
English-Chinese Books  & En $\Leftrightarrow$ Zh & paragraph-level & $60,000$ \\
Yiyan Corpus  & En $\Leftrightarrow$ Zh & sentence-level & $30,000$ \\
United Nations Parallel Corpus  & En $\Rightarrow$ Zh/ Fr/ Es/ Ru/ Ar & sentence-level & $60,000$\\
\bottomrule
\end{tabular}
}
\caption{\label{mt-datasets}
Details of translation training data. In English-Chinese Books dataset and Yiyan Corpus dataset, we simultaneously use both directions of parallel corpora. In United Nations Parallel Corpus, we utilize approximately $60,000$ samples from English to each language.
}
\end{table*}

    \begin{figure}[t]
        \centering
        \includegraphics[width=1.05\columnwidth]{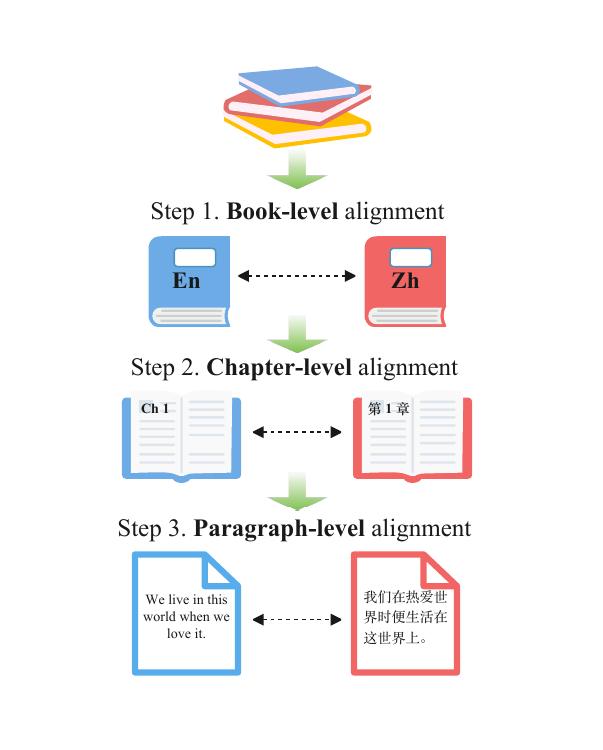}
        \caption{The process of constructing the English-Chinese book dataset.}
        \label{fig:book}
    \end{figure}
\noindent\textbf{English-Chinese Books.} In order to collect rich human expression habits in book translation data, we manually construct an English-Chinese parallel book corpus dataset. 
The construction process of this dataset, as shown in Figure ~\ref{fig:book}, can be divided into three steps:
Firstly, alignment at the \textbf{book level}. We manually collect Chinese and English versions of several books, ensuring high quality for both versions selected, with translations being provided by skilled professional translators.
Next, alignment at the \textbf{chapter level} is performed for each book's Chinese and English versions. We parse the data of the entire book into text format and then compare the number and content of chapters for consistency.
Finally, we align Chinese and English paragraphs at the \textbf{paragraph level} for each chapter through manual comparison and adjustment.

\noindent\textbf{Yiyan Corpus.}\footnote{https://corpus.bfsu.edu.cn/info/1070/1631.htm} To enhance the diversity of the data and strengthen the model's robustness to inputs of different lengths, we incorporate the Yiyan corpus, an English-Chinese Parallel Corpus. Specifically, we utilize the academic and novel sections, consisting of parallel sentences translated by human translators at the sentence level.

\noindent\textbf{United Nations Parallel Corpus (UN).} \cite{UN_dataset}
For our multilingual experiments, we use the UN training set, which was also manually translated. 
This dataset includes parallel data in six languages: English, Chinese, French, Spanish, Russian, and Arabic. We conduct experiments on translation from English to the other five languages.
We randomly sample from the extensive dataset, ensuring English sentences contain a minimum of 30 words to guarantee richer information. 

In the experiment for bidirectional English-Chinese translation, we mix English-Chinese books data with Yiyan Corpus data. For the multilingual experiment, we utilize the UN dataset.

\subsection{Model}
\begin{itemize}
  \item Ultra-LLaMA2-7B: Base model of our experiments. A variant of LLaMA2-7B further-pretrained on over $200B$ Chinese tokens.
  \item LLaMA2-7B ~\cite{llama2}: A LLM trained primarily in English. In certain experiment, we use this model as the control.

\end{itemize}
\subsection{Evaluation}
\subsubsection{Metrics}
When evaluating the quality of translation results, we employed three evaluation methods: GPT-4 comparative evaluation ~\cite{GPT4} and COMET metrics ~\cite{COMET} and human evaluation.

    \begin{figure}[t]
        \centering
        \includegraphics[width=\columnwidth]{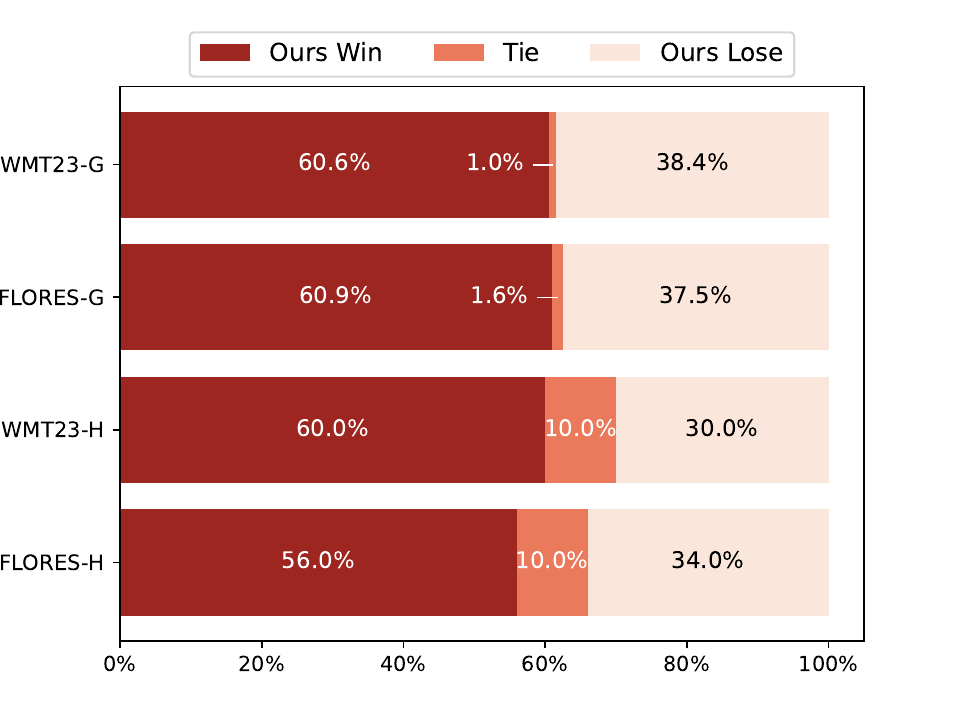}
        \caption{Comparison between preference optimized models and the SFT model on Task En$\rightarrow$Zh. G and H represent GPT-4 and humans as evaluators, respectively.}
        \label{fig:ultrallama2-enzh}
    \end{figure}

    \begin{figure}[t]
        \centering
        \includegraphics[width=\columnwidth]{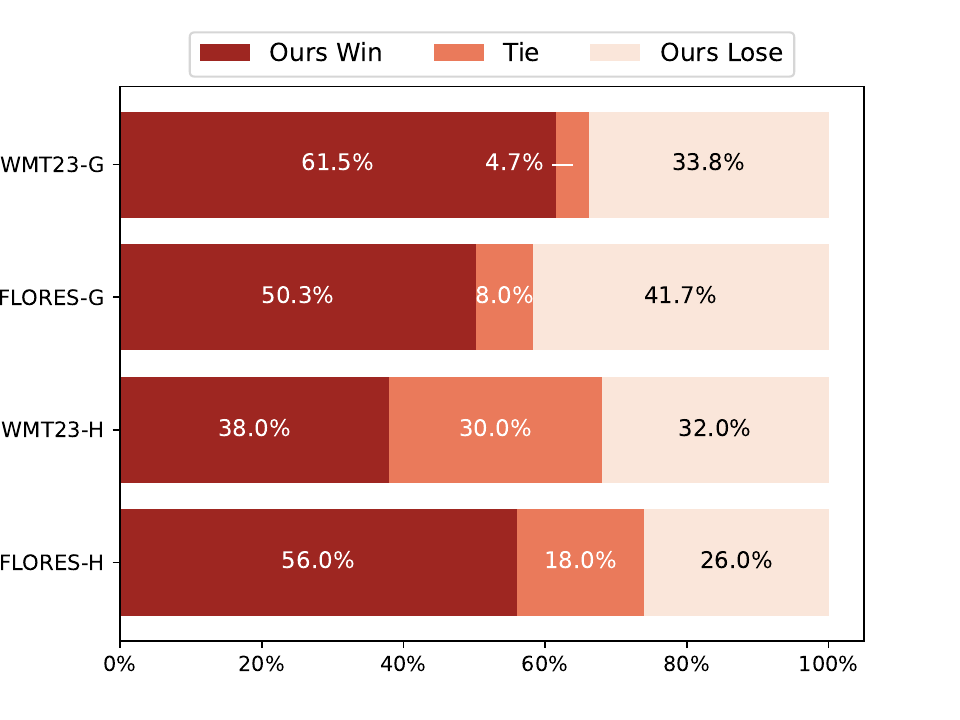}
        \caption{Comparison between preference optimized models and the SFT model on Task Zh$\rightarrow$En. G and H represent GPT-4 and humans as evaluators, respectively.}
        \label{fig:ultrallama2-zhen}
    \end{figure}
    
\noindent\textbf{GPT-4.} Due to its exceptional general-purpose capabilities, the GPT-4 model has emerged as a pioneering approach for evaluating NLP tasks. We present the original text of a given sentence alongside translations from both the SFT and RLHF models, allowing GPT-4 to compare them simultaneously and select the superior translation. In the prompt used during the tests, we explicitly included multidimensional evaluation criteria, including flexibility, fidelity, and accuracy and so on. To mitigate the impact of comparison order, we interchanged the positions of both models' outputs for each test, conducting two evaluations simultaneously. Refer to the Table ~\ref{prompt_template} in appendix for the complete prompt.

\noindent\textbf{COMET.} COMET is a neural framework for training multilingual machine translation evaluation models. It has been shown to have high correlation with human assessment and has become an increasingly widely used metric for machine translation evaluation ~\cite{DBLP:conf/wmt/KocmiFGJMM21}. We select the reference-free quality evaluation model wmt22-cometkiwi-da ~\citet{kiwi}.
We compare the translation abilities of two models (SFT and RLHF models) by evaluating the relative COMET scores of their translation results for the same translated data.

\noindent\textbf{Human Evaluation.} When evaluating bidirectional English-Chinese translation, we also incorporate human evaluation. 
Proficient bilingual native speakers conduct assessments to compare translation quality.


\subsubsection{Test Sets}
We utilize the WMT23 test sets ~\cite{WMT23} and the Flores-200 devtest sets ~\cite{flores-200} to assess the model's performance. Note that WMT23 does not cover all directions for the multilingual experiment, but as we employ comparative reference-free evaluation, we only use English data from the WMT23 test sets as the source.

\section{Results and Disscussions}
\subsection{Main Results}

\begin{table*}[h]
    {  \footnotesize \begin{tabularx}{\linewidth}{cc X} 
    \toprule 
    \multirow{8}{*}{\textbf{Faithfulness}} &Input  & {The synthesis of \colorbox {yellow}{the pharmaceutical compound acetylsalicylic acid}, commonly known as aspirin, marked a significant advancement in modern medicine.}\\
    &SFT  & {\begin{CJK}{UTF8}{gbsn} 阿司匹林的合成标志着现代医学的一个重要进步。\end{CJK}}\\
    &RLHF  & {\begin{CJK}{UTF8}{gbsn} \colorbox {yellow}{乙酰水杨酸}（阿司匹林）\colorbox {yellow}{这种药物}的合成，标志着现代医学的一个重要进步。\end{CJK}}\\
    &Commentary  & {In the translation by RLHF, the term `\begin{CJK}{UTF8}{gbsn}乙酰水杨酸这种药物\end{CJK}' corresponds to `the pharmaceutical compound acetylsalicylic acid' in the input text, while in the translation by SFT, this expression is missing, reflecting an improvement in translation faithfulness.}\\
    \hline\hline
    \multirow{7}{*}{\textbf{Expressiveness}} &Input  & {After years of practice, running a marathon was \colorbox {yellow}{a piece of cake} for her.}\\
    &SFT  & {\begin{CJK}{UTF8}{gbsn} 经过多年的练习，对她来说，跑马拉松就\colorbox {yellow}{像吃蛋糕一样简单}。\end{CJK}}\\
    &RLHF  & {\begin{CJK}{UTF8}{gbsn} 经过多年的锻炼，跑马拉松对她来说已是\colorbox {yellow}{小菜一碟}了。\end{CJK}}\\
    &Commentary  & {In the SFT translation, `\begin{CJK}{UTF8}{gbsn}像吃蛋糕一样简单\end{CJK}' is a literal translation of "a piece of cake" in the input text. In contrast, the translation in RLHF, `\begin{CJK}{UTF8}{gbsn}小菜一碟\end{CJK}', is a more authentic Chinese expression, vivid and expressive. This case reflecting an enhancement in the expressive power of the translation.}\\
    \hline\hline
    \multirow{10}{*}{\textbf{Elegance}} &Input  & {As the crimson hues of dusk melded with the cerulean tapestry of the night sky, the poet pondered over verses that could encapsulate the \colorbox {yellow}{ephemeral} beauty of the twilight.}\\
    &SFT  & {\begin{CJK}{UTF8}{gbsn} 夜幕降临，天空中的蓝色帷幕与黄昏的红色调和在一起，诗人开始思考如何用诗句来捕捉这\colorbox {yellow}{短暂}的美好。\end{CJK}}\\
    &RLHF  & {\begin{CJK}{UTF8}{gbsn} 暮色渐浓，绯红的余晖与夜空的青蓝交织，诗人思忖着如何用诗句来捕捉这\colorbox {yellow}{转瞬即逝}的美景。\end{CJK}}\\
    &Commentary  & {Both `\begin{CJK}{UTF8}{gbsn}转瞬即逝\end{CJK}' and `\begin{CJK}{UTF8}{gbsn}短暂\end{CJK}' can be used to convey the meaning of `ephemeral' in the input text, but the former implies a sense of regret and sorrow for the fleeting nature of beautiful things, while the latter is a neutral term, simply describing temporal brevity. This example demonstrates an improvement in the elegance of the translation.}\\
  \bottomrule 
 \end{tabularx} }
    \caption{An case study on modeling human translation preference through RLHF. The yellow background text reflects the improved translation quality of RLHF compared to SFT.}
    \label{tab:case}
\end{table*}
\noindent \textbf{Is it feasible to model translation preferences without explicit preference annotations?}

\noindent This paper explores the feasibility of modeling human translation preferences in the absence of explicit preference annotations. By comparing the deficiencies of machine translation with human translation, the reward model learns human translation preferences, thus circumventing the need for costly preference data annotation. In this subsection, we empirically validate the effectiveness of this approach. Specifically, we use high-quality English-Chinese parallel corpora (refer to Section \ref{sec:data_collection}) as preferred data, while data generated by the SFT model (also fine-tuned using pre-heldout book data) serves as dispreferred data. From Figure \ref{fig:ultrallama2-enzh} and \ref{fig:ultrallama2-zhen}, we observe that on the WMT23 and FLORES datasets, our preference-optimized model exhibits significantly improved win rates compared to the SFT model, regardless of whether the evaluator is GPT-4 or human. This indicates that with access to high-quality parallel corpora, even in the absence of explicit preference annotations, we can learn human translation preferences and improve the translation quality of the model. In Table \ref{tab:case}, we demonstrate the quality improvement of translations after preference optimization through three cases.

    \begin{figure}[t]
        \centering
        \includegraphics[width=\columnwidth]{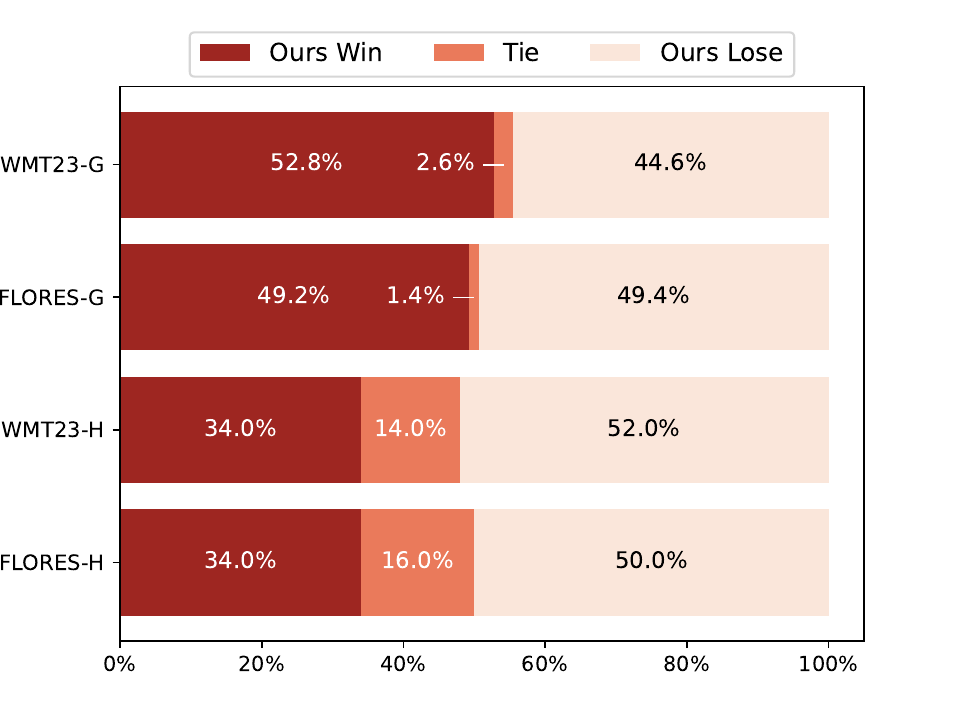}
        \caption{After replacing the base model in Figure \ref{fig:ultrallama2-enzh} with LLaMA, compare the preference optimized model and the SFT model in the En$\rightarrow$Zh translation direction.}
        \label{fig:llama2-zhen}
    \end{figure}

        \begin{table*}
            \centering
            \begin{tabular}{cc ccc ccc}
            \toprule
            \multirow{2}{*}{\textbf{Dataset}} & \multirow{2}{*}{\textbf{Evaluator}}& \multirow{2}{*}{\textbf{Results}} & \multicolumn{5}{c}{Translation Direction}\\
            \cline{4-8}
            &&& En$\rightarrow$Fr & En$\rightarrow$Es & En$\rightarrow$Ru & En$\rightarrow$Zh & En$\rightarrow$Ar
            \\
            \midrule
            \multirow{6}{*}{WMT23}
            & \multirow{3}{*}{GPT-4} & SFT Win & $\bm{0.510}$ & $0.432$ & $0.462$ & $0.395$ & $0.447$\\
            &  & RLHF Win & $0.430$ & $\bm{0.439}$ & $\bm{0.490}$ & $\bm{0.552}$ & $\bm{0.534}$\\
            &  & Tie & $0.060$ & $0.129$ & $0.048$ & $0.053$ & $0.019$\\
            \cline{2-8}
            &\multirow{3}{*}{COMET}& SFT Win & $0.416$ & $0.386$ & $0.450$ & $0.326$ & $0.450$\\
            & & RLHF Win & $\bm{0.544}$ & $\bm{0.506}$ & $\bm{0.516}$ & $\bm{0.634}$ & $\bm{0.550}$\\
            &  & Tie & $0.040$ & $0.108$ & $0.034$ & $0.040$ & $0.000$\\
            \hline\hline
            \multirow{6}{*}{FLORES}
            & \multirow{3}{*}{GPT-4} & SFT Win & $\bm{0.495}$ & $0.378$ & $0.455$ & $0.347$ & $0.416$\\
            &  & RLHF Win & $0.417$ & $\bm{0.396}$ & $\bm{0.477}$ & $\bm{0.587}$ & $\bm{0.552}$\\
            &  & Tie & $0.088$ & $0.226$ & $0.068$ & $0.066$ & $0.032$\\
            \cline{2-8}
            &\multirow{3}{*}{COMET}& SFT Win & $0.398$ & $0.344$ & $0.424$ & $0.328$ & $0.448$\\
            & & RLHF Win & $\bm{0.536}$ & $\bm{0.472}$ & $\bm{0.526}$ & $\bm{0.624}$ & $\bm{0.552}$\\
            &  & Tie & $0.066$ & $0.184$ & $0.050$ & $0.048$ & $0.000$\\
            \bottomrule
            \end{tabular}
            \caption{Results of preference modeling in five translation directions on the UN dataset.}
            \label{tab:multi_lang}
        \end{table*}    
    
\noindent\textbf{The language capability of reward model is crucial for preference learning.}

\noindent In the previous part of the experiment, we utilize Ultra-LLaMA as the base model, which is a variant of LLaMA further-pretrained on over $200B$ Chinese tokens. To investigate the impact of language capability differences on preference learning, we replace the base model with original LLaMA, which has a relatively weaker processing capability for Chinese. We construct the SFT model using the same experimental data and training scheme as in the previous section and further optimize it for human preferences. As observed from Figure \ref{fig:llama2-zhen}, the win rate of the preference-optimized model significantly decreased in comparison with the SFT model, and it even lost to the SFT model in human evaluations. It is worth noting that the SFT model trained on original LLaMA inherently lacks translation capabilities compared to the SFT model based on Ultra-LLaMA, thus highlighting more pronounced differences in the quality of generated translations compared to human translations. Intuitively, this should decrease the learning difficulty of the reward model. However, the reward model constructed based on original LLaMA failed to effectively model human translation preferences. Therefore, we believe that the language capability of reward models plays an important role in preference learning.
        \begin{table*}
            \centering
            \resizebox{\linewidth}{!}{
            \begin{tabular}{cc ccc ccc}
            \toprule
            \multirow{2}{*}{\makecell[c]{\textbf{Translation Direction}\\ \textbf{Optimized by RLHF}}} & \multirow{2}{*}{\textbf{Evaluator}}& \multirow{2}{*}{\textbf{Results}} & \multicolumn{5}{c}{Transferred Translation Direction}\\
            \cline{4-8}
            &&& En$\rightarrow$Fr & En$\rightarrow$Es & En$\rightarrow$Ru & En$\rightarrow$Zh & En$\rightarrow$Ar
            \\
            \midrule
            \multirow{6}{*}{En$\rightarrow$Zh}
            & \multirow{3}{*}{GPT-4} & SFT Win & $0.443$ & $0.448$ & $0.418$ & $-$ & $0.355$\\
            &  & RLHF Win & $\bm{0.540}$ & $\bm{0.493}$ & $\bm{0.563}$ & $-$ & $\bm{0.563}$\\
            &  & Tie & $0.018$ & $0.030$ & $0.020$ & $-$ & $0.083$\\
            \cline{2-8}
            &\multirow{3}{*}{COMET}& SFT Win & $0.390$ & $0.410$ & $0.475$ & $-$ & $0.420$\\
            & & RLHF Win & $\bm{0.610}$ & $\bm{0.590}$ & $\bm{0.525}$ & $-$ & $\bm{0.580}$\\
            &  & Tie & $0.000$ & $0.000$ & $0.000$ & $-$ & $0.000$\\
            \hline\hline
            \multirow{6}{*}{En$\rightarrow$Ar}
            & \multirow{3}{*}{GPT-4} & SFT Win & $0.458$ & $\bm{0.465}$ & $0.455$ & $\bm{0.485}$ & $-$\\
            &  & RLHF Win & $\bm{0.510}$ & $0.458$ & $\bm{0.533}$ & $\bm{0.485}$ & $-$\\
            &  & Tie & $0.033$ & $0.078$ & $0.013$ & $0.030$ & $-$\\
            \cline{2-8}
            &\multirow{3}{*}{COMET}& SFT Win & $0.410$ & $\bm{0.505}$ & $0.435$ & $\bm{0.580}$ & $-$\\
            & & RLHF Win & $\bm{0.590}$ & $0.495$ & $\bm{0.565}$ & $0.420$ & $-$\\
            &  & Tie & $0.000$ & $0.000$ & $0.000$ & $0.000$ & $-$\\
            \bottomrule
            \end{tabular}
            }
            \caption{Cross-lingual Transfer Results of Translation Preferences.}
            \label{tab:transfer}
        \end{table*}    
    
\subsection{The Impact of the Inherent Nature of Human Translation}
    \begin{figure}[t]
        \centering
        \includegraphics[width=\columnwidth]{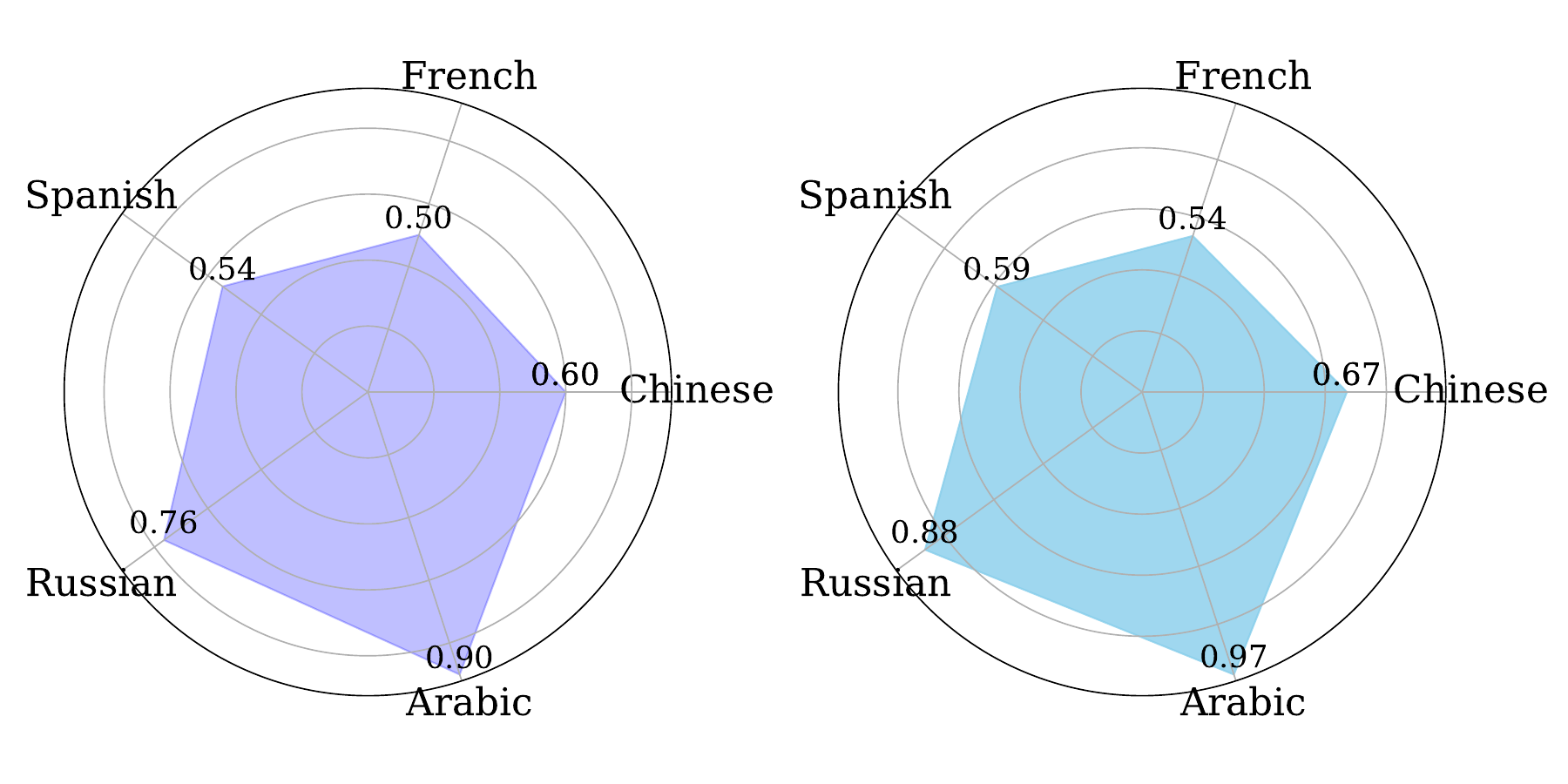}
        \caption{Quality Analysis of UN Datasets.}
        \label{fig:comet}
    \end{figure}

The book dataset used in the previous section has high textual quality, containing complex linguistic structures and grammar phenomena, and is diverse in its domain sources. In contrast, the UN originates from specific domains and lacks complex linguistic structures and rhetorical devices commonly found in governmental documents. 
In this section, we conduct multilingual experiments using the UN dataset to explore the influence of intrinsic properties of the data on preference learning.

\noindent\textbf{For simple domain-specific parallel corpora, the quality of machine translations is comparable to human translations.}

\noindent As shown in Figure \ref{fig:comet} (left), using COMET as the evaluation metric, we find that the difference in quality between translations from the SFT model and human translations is minimal. Especially for French and Spanish, only $50\%$ and $54\%$ of human translations respectively outperform translations from the SFT model. This indicates that when parallel corpora do not contain complex linguistic sources or sentence structures, the SFT model can already achieve results comparable to human translations. Clearly, the induction bias of "human translations are superior to translations from the SFT model" is no longer valid for such datasets.

\noindent\textbf{Similar translation quality increases the difficulty of preference learning.}

\noindent To explore preference learning on the United Nations dataset, we first remove $50\%$ of the data with small differences in COMET scores, retaining data pairs with relatively clear preference tendencies. However, as shown in Figure \ref{fig:comet} (right), in the directions of French and Spanish, nearly $50\%$ of SFT translations still outperform human translations. Therefore, we reannotate based on COMET scores to construct a preference dataset. As shown in Table \ref{tab:multi_lang}, translation models optimized for preferences significantly outperform the SFT model in all five translation directions in terms of COMET scores. This is easily understood since our preference labels are derived from COMET scores. However, learned preferences may not necessarily be generalizable and aligned with human preferences. The evaluation results of GPT-4 in Table \ref{tab:multi_lang} indicate that in the English to Spanish and Russian directions, the preference-optimized model only has a slight advantage, and in the case of French, it even loses to the SFT model. This is mainly because the difference between SFT and human translations is minimal in French. In contrast, in the English to Arabic direction, the preference-optimized model consistently and significantly improves, mainly due to the distinct differences in preference data itself, making it easier for the reward model to learn generalizable translation preferences.

\subsection{Transferability Analysis}
With the powerful Chinese capabilities of the reward model and the notable quality disparities in Arabic preference data, translation models have achieved effective alignment with human preferences in both English-to-Chinese and English-to-Arabic directions. In this section, we explore through experiments whether learned translation preferences can be transferred across languages. As observed from Table \ref{tab:transfer}, RLHF training solely on tasks in English-to-Chinese translation, the learned human preferences can effectively transfer to other languages and consistently improve performance. Similarly, when English-to-Arabic translation is used as the source task, improvements are also evident in tasks such as English-to-French and English-to-Russian translation. This indicates that aligning with and transferring from human preferences in other translation directions can be a viable strategy when the current translation direction lacks reward models with strong language capabilities or high-quality preference data.

\section{Conclusions}
This paper explores modeling translation preferences with RLHF to improve the quality of machine translation. We propose a cost-effective preference learning strategy, optimizing reward models by contrasting deficiencies in machine translation compared to human translation. Learning human preferences while avoiding expensive preference data annotation. Further analysis suggests that the language capability of the reward model and the nature of the data itself affect the effectiveness of preference learning. Additionally, learned preferences exhibit cross-lingual transfer phenomena. This may be beneficial for preference modeling in low-resource languages.

\section*{Limitations}
Due to cost limitations, we only collected English-Chinese aligned book data as a substitute for preference data, without covering more translation directions. Additionally, our human evaluations were limited to English-Chinese translation, with GPT-4 used as a proxy for manual evaluations in other translation directions. In the future, we will attempt to align with human translation preferences in more languages, especially low-resource languages, and conduct comprehensive manual evaluations in more translation directions.

\bibliography{anthology,custom}
\bibliographystyle{acl_natbib}

\appendix
\section{Implementation Details}

\noindent\textbf{SFT stage.}
In the English-Chinese model, we use $1/3$ of the dataset, with a learning rate of $5e-6$, training for $2$ epochs; In the multilingual model, approximately $3/4$ of the training data is used for $1$ epoch, with a learning rate of $5e-6$.

\noindent\textbf{RM training stage.}
The reward model is initialized with the previous stage's SFT model. In the English-Chinese model, the remaining $2/3$ of the training data are used to form chosen-rejected pairs with the data generated by the SFT model; In the multilingual model, the remaining $1/4$ of the training data is utilized, and only the top $50\%$ of high-confidence data selected by the COMET model, is used to train the RM. Training continues with dynamic batch processing until early stopping criteria are met.

\noindent\textbf{RL stage.}
For English-Chinese model, we reuse the inputs from the RM stage's training data as queries, and for multilingual model, we use English monolingual book data obtained from web crawling as queries. We set the KL divergence penalty coefficient to $0.02$, and trained until early stopping criteria were met. 
\begin{table}[!ht]
\centering
\begin{tabular}{p{\linewidth}}
\toprule
You are a translation expert, and I need your help in impartially judging the quality of two translations. The judging criteria are as follows:\\
Flexibility of Translation: A good translation is not confined to the original form, and it should be smooth and clear. Poor-quality translations appear rigid and awkward, merely translating word-for-word according to the original form.\\
Fidelity of Translation: A good translation should faithfully reflect the content of the original text. It should not introduce content that does not exist in the original, nor should it omit content present in the original.\\
Accuracy and Elegance of Phrasing: In a good translation, phrases and wording should adhere to the conventions of the target language, and they should be as accurate and elegant as possible.\\
Next, I will provide you with the original text and two translations. Please let me know which one is better according to these criteria. Please give your judgment directly and do not output additional explanations.\\
\bottomrule
\end{tabular}
\caption{Prompt template for GPT4 evaluaiton.}
\label{prompt_template}
\end{table}\\

\end{document}